\pdfoutput=1

\PassOptionsToPackage{table,dvipsnames,prologue}{xcolor}

\documentclass[11pt]{article}

\usepackage[]{acl}

\usepackage{times}
\usepackage{latexsym}
\newcommand{\dbcomment}[1]{\textcolor{blue}{[\textsc{#1} -- DB]}}
\newcommand{\ignore}[1]{}
\usepackage[T1]{fontenc}

\usepackage[utf8]{inputenc}

\usepackage{microtype}
\usepackage{tikz}
\usepackage{pgfplots}
\pgfplotsset{compat=1.18}
\usepackage{amsfonts}

\usepackage{amsmath}
\usepackage{booktabs}
\usepackage{subcaption}
\usepackage{bbm}
\usepackage{fbox}

\definecolor{myorange}{RGB}{230, 145, 56}
\definecolor{myblue}{RGB}{60, 12, 216}

\usepackage{array}
\newcolumntype{H}{>{\setbox0=\hbox\bgroup}c<{\egroup}@{}}

\title{Social Meme-ing: Measuring Linguistic Variation in Memes}

\author{Naitian Zhou,$^1$ David Jurgens$^2$\and David Bamman$^1$ \\
  $^1$University of California, Berkeley \\
  $^2$University of Michigan \\
  \texttt{\{naitian,dbamman\}@berkeley.edu}, 
  \texttt{jurgens@umich.edu} \\}

\begin{document}
\maketitle
\begin{abstract}
Much work in the space of NLP has used computational methods to explore sociolinguistic variation in text.  In this paper, we argue that memes, as multimodal forms of language comprised of visual templates and text, also exhibit meaningful social variation.
We construct a computational pipeline to cluster individual instances of memes into templates and semantic variables, taking advantage of their multimodal structure in doing so.
We apply this method to a large collection of meme images  from Reddit and make available the resulting \textsc{SemanticMemes} dataset of 3.8M images clustered by their semantic function. We use these clusters to analyze linguistic variation in memes, discovering  not only that socially meaningful variation in meme usage exists between subreddits, but that patterns of meme innovation and acculturation within these communities align with previous findings on written language.
\end{abstract}

\section{Introduction}

One objective in variationist sociolinguistics is to study how social factors contribute to differences in the way people use language. Work in natural language processing has followed this tradition, offering large-scale analyses of how language use is conditioned on geography, \citep{eisenstein-etal-2010-latent,hovy-purschke-2018-capturing,demszky-etal-2021-learning},  community\ \citep{del-tredici-fernandez-2017-semantic,zhu2021structure,lucy-bamman-2021-characterizing} and time\ \citep{hamilton-etal-2016-diachronic}.
This work is important not only because language variation often exposes shortcomings in NLP tools, which are primarily developed for standard language varieties \citep{blodgett-etal-2016-demographic}, but also because variation often embeds \textbf{social meaning}. We make inferences about people's social class, regionality, gender, and much more based on the way they talk \cite{campbell-kiblerNatureSociolinguisticPerception2009, zhangChineseYuppieBeijing2005}, and we strategically use language to actively construct and perform identities \cite{labovSocialMotivationSound1963a, bucholtzIdentityInteractionSociocultural2005}.

\begin{figure}
    \centering
    \includegraphics[width=.9\linewidth]{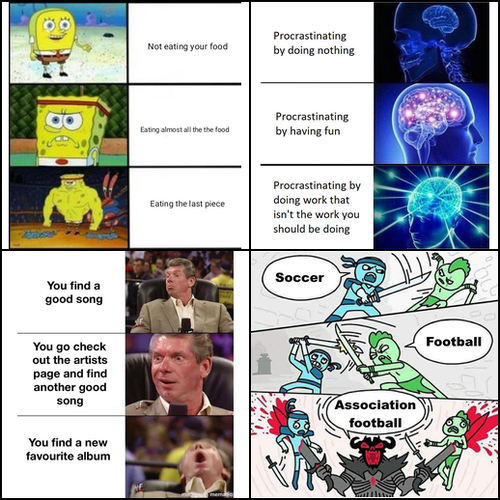}
    \caption{Meme templates can be visually diverse, but often provide the same semantic function; in this case, all four templates show a scalar increase.
    }
    \label{fig:intro}
\end{figure}

Most of this work has focused on lexical or morphosyntactic variation in written texts. However, language exists beyond text or speech. In face-to-face interaction, multimodality in language has been construed as features like co-speech gesture, facial expression or body movement \cite{pernissWhyWeShould2018}. In online communication, previous work has extended the term to include the interplay between images and text \citep{kress2001, zhang-etal-2021-multimet, hessel-etal-2023-androids}. Understanding text in isolation is insufficient to understanding how we communicate online.%

\begin{figure}
    \centering
    \includegraphics[width=0.8\linewidth]{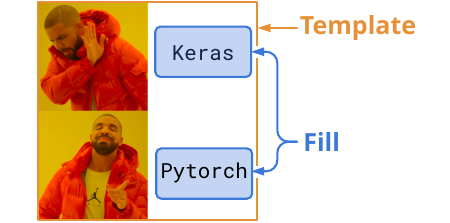}
    \caption{Memes are multimodal constructions where the base image \textbf{\textcolor[HTML]{E69138}{template}} and additional text \textbf{\textcolor[HTML]{3C78D8}{fills}} both have semantic value.}
    \label{fig:decompose}
\end{figure}

In the space of multimodal online language, memes are interesting for their compositionality. They consist of a base image (the \textbf{template}) as well as superimposed text (which we refer to as the \textbf{fill}). For example, the ``Drake'' template depicted in figure~\ref{fig:decompose} serves the semantic function of expressing a preference relation between the fills.  This same Drake template can be used to express preference relations between a range of fills; at the same time,
 multiple different templates can share the same or similar semantic function, as illustrated in fig. \ref{fig:intro} for the function of ``scalar increase.''
We refer to this set of functionally equivalent templates as a semantic \textbf{cluster}.

In this work, we follow the variationist sociolinguistics tradition by treating templates as \emph{variants} and semantic clusters as \emph{variables}, observing how social factors might contribute to the distribution among these variants. To conduct this analysis, we develop a method for identifying semantic clusters by exploiting the visual structure of meme templates and the linguistic structure of meme fills. We use this to create the \textsc{SemanticMemes} dataset of 3.8M Reddit memes\footnote{We make data and code available at \url{https://github.com/naitian/semantic-memes}} grouped into semantic clusters and validated with a human evaluation. Finally, we use these semantic clusters to perform a series of case studies demonstrating their use in studying linguistic variation, linguistic innovation, and linguistic acculturation. We find that:

\begin{enumerate}
    \item socially meaningful variation in template choice exists between subreddits; \item subreddits that first introduce a new template continue to use it more than others; and 
    \item users who stay in a subreddit for longer tend to use templates distinctive to that subreddit. 
    \end{enumerate}
These findings illustrate the ways in which memes function as multimodal acts of communication, and how methods from computational sociolinguistics can shed light on meaningful  variation within them.

\ignore{
\dbcomment{Parts of this could maybe be sprinkled in}
Memes on Reddit are remarkably natural candidates for such an analysis, for the following reasons:
\begin{enumerate}
    \item \textbf{Discreteness} Meme templates can be reasonably construed as discrete items in a multimodal lexicon. Though memes and meme templates are often remixed \cite{wigginsMemesGenreStructurational2015}, they mostly retain their structure and semantic function. This allows us to naturally apply to meme templates the same analysis used for other discrete lexical items (namely, words).

    \item \textbf{Separation of content and style} We also observe that there exists great visual diversity in memes that are functionally similar. One challenge in computationally modeling social meaning is the difficulty in separating it from referential meaning. Because memes exist as a mix of modalities, we can represent meme content as a function of one modality (the fill text) while observing the variation in the meme style isolated to the other modality (the template image).

    \item \textbf{Reddit communities of practice} Reddit as a social media platform also lends itself to sociolinguistic analysis because it is organized into subreddits. These form communities of practice where users learn the language and values necessary to engage with each other. Language is used not only to engage in discourse within the community, but to identify with it \cite{holmesCommunityPracticeTheories1999}.

\end{enumerate}
}

\section{Methods}

\begin{figure*}[t]
    \centering
    \includegraphics[width=0.8\textwidth]{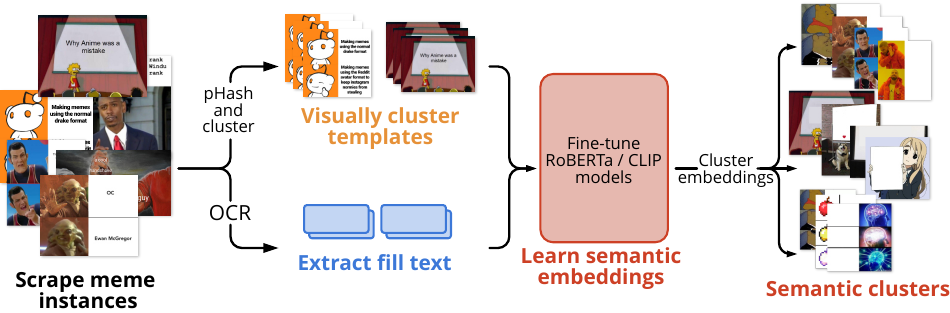}
    \caption{We group visually identical meme instances into templates, and extract the fills using OCR. This data is used to learn semantic embedding representations of templates, which we use to generate semantic clusters.}
    \label{fig:pipeline}
\end{figure*}

To study variation in meme use, we need to identify the meme variables that organize a collection of meme \textbf{instances}---the individual memes that are created and posted online by specific people at specific moments in time. 
We create a pipeline that visually clusters meme \textbf{instances} into \textbf{templates} (i.e., the same memes that differ by variation in fills) by exploiting the visual similarity between them; and linguistically clustering meme \textbf{templates} into \textbf{semantic clusters} by exploiting the similarity among the fills used in different templates.  Fig. \ref{fig:pipeline} provides an overview of the process, which involves first clustering instances into templates (\S\ref{templates}), and then clustering templates into variables (\S\ref{langclustering}).

\subsection{Visually clustering instances}
\label{templates}

Our process starts with a set of meme instances, which we wish to group based on visual similarity; this process serves to group memes into their base templates as well as filter out many non-meme images. This is difficult due to the massive number of images as well as the amount of variation in zoom, crop, borders and other visual details. We lay out the steps of the process here, but provide further details and example images in Appendix~\ref{sec:visual_details}.

We first preprocess images to remove any solid color framing elements to isolate the base image, then follow\ \citet{Zannettou2018} and \ \citet{Morina2022} in extracting templatized memes by running a perceptual hashing algorithm.%

We then compute the pairwise Hamming distance between hashes that occur more than 10 times, discard any pairs where the distance was greater than a cut-off $d_{\text{max}} = 10$. We use the Leiden clustering algorithm to perform clustering \ \cite{traagLouvainLeidenGuaranteeing2019}.\footnote{We found that using DBSCAN, as was done in prior work, resulted in many images being put into a single noisy cluster.} The Leiden algorithm iteratively finds well-connected subgraphs; we construct a graph where image hashes were vertices and the edge weight was $e_{ij} = d_{\text{max}} - d_{ij} + 1 $ for vertices $i$ and $j$, where $d_{ij}$ was the Hamming distance between them.

The clustering algorithm splits aggressively---instances with similar base images may be split across multiple templates due to variations in the zoom, crop, and borders. We find the next step, which clusters based on the fill text, serves as a remedy by placing many of these duplicate templates into the same semantic cluster.
Appendix~\ref{sec:visual_details} contains examples of template clusters.

\subsection{Linguistically clustering templates}\label{langclustering}

Given a set of meme templates, we want to identify clusters of those templates that have a similar semantic function---i.e., that are used to assert a similar relation among the text in the fills (such as a comparison function exemplified by the Drake meme in fig. \ref{fig:decompose}). These semantic clusters are the linguistic variables of analysis: discrete sets of variants which share a semantic function but vary in the social meanings they index.

We apply the key intuition that people will use certain templates to make certain classes of statements (comparison, declaration, surprise); as with any other language, fills that are ``grammatical'' for one template may be nonsensical in another.  Templates that share similar sets of fills, then, may perform a similar function over them.

To cluster templates using this principle, we extract the fill text from meme instances belonging to a template (\S\ref{fills}), learn semantic representations for templates based on the distribution of text fills (\S\ref{semantic}), and cluster those representations (\S\ref{variables}).

\subsubsection{Extracting fill text}\label{fills}

We extract text (along with the bounding boxes containing it) from meme instances using EasyOCR.\footnote{https://github.com/JaidedAI/EasyOCR} We use the order of the bounding boxes as a rough signal for the position and ordering of the text, but do not incorporate the bounding coordinates directly into the models described below.

Some meme templates contain text in the base image. To prevent these from trivializing the semantic embedding task, we remove bounding boxes with text that was identical in over 90\% of the memes in a given template cluster.%

\subsubsection{Learning semantic embeddings}\label{semantic}

We examine four methods for learning semantic embeddings of memes, each described in more detail below: a RoBERTa classifier fine-tuned to predict the template given the fill text; a CLIP model fined-tuned on (fill text, image) pairs; the vector difference between fine-tuned and pretrained CLIP embeddings (CLIP-diff); and concatenating  CLIP-diff  and RoBERTa embeddings (Concat).

\paragraph{Text-only RoBERTa.}

In the text-only model, we fine-tune a RoBERTa model on a sequence classification task to predict a distribution over templates given the fill text as input. We separate text in different bounding boxes in a meme with a separator token when passing it into the model to impose a rough, linear notion of space. %

After fine-tuning, we take the weights of the final classification layer $W \in \mathbb{R}^{768 \times N}$ as the embeddings, where $N$ is the number of templates. Intuitively, RoBERTa is an encoder model that projects the fill text into a latent semantic space. The final classification layer can be thought of as a projection from that latent space into the discrete space of templates. Therefore, the transposition of the weight matrix can be viewed as a mapping from templates into the latent semantic space.

\begin{figure*}
    \centering

    \begin{subfigure}{0.49\textwidth}
    \includegraphics[width=\textwidth]{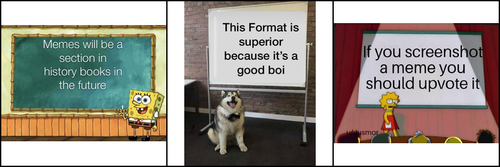}
    \caption{Declaration}
    \end{subfigure}%
    ~
    \begin{subfigure}{0.49\textwidth}
    \includegraphics[width=\textwidth]{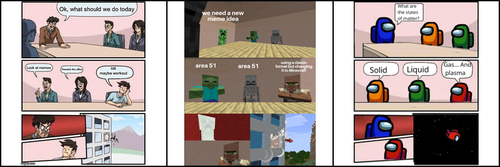}
    \caption{Unpopular statement}        
    \end{subfigure}%
    
    \begin{subfigure}{0.49\textwidth}
    \includegraphics[width=\textwidth]{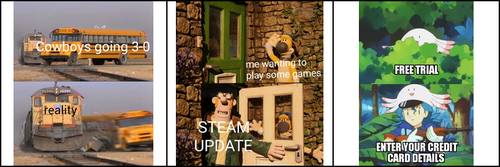}
    \caption{Surprise narrative}
    \end{subfigure}%
    ~
    \begin{subfigure}{0.49\textwidth}
    \includegraphics[width=1\linewidth]{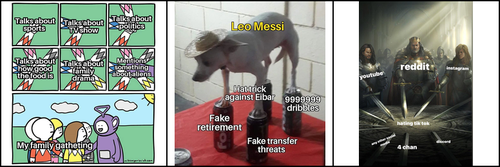}
    \caption{Similarity / parts of a whole}
    \end{subfigure}%

    \caption{Examples from semantic clusters generated from RoBERTa embeddings; visually diverse clusters emerge even for complex semantic functions like a surprise narrative.}
    \label{fig:cluster-examples}

\end{figure*}

\paragraph{Multimodal CLIP.}

In learning the embeddings, the text-based RoBERTa model does not have direct access to the image features in the templates. We experiment with using both the image and text data by fine-tuning a CLIP model.%

We fine-tune CLIP using a contrastive loss between the embedding of a meme instance and its fill text. To prevent the model from cheating by reading the text in the image, we sample a meme instance with different text but the same template. This fine-tuning step modifies the image embedding to align with fill text, which implicitly describes the semantic function of the meme, instead of with the pretraining dataset of image captions, which explicitly describe the contents of the image.

CLIP generates embeddings of meme instances. To generate template embeddings, we randomly sample up to ten instances of a template as input for the image embedding module. We then compute the average image embedding of those instances. We don't embed the fill text for this step, since fill text greatly varies between meme instances that use the same template, but the image templates should be more or less visually identical. %

\paragraph{CLIP-diff.}

It is possible that the fine-tuned model contains a notion of meme semantics that is in tension with the pretraining task of image captioning. To isolate the meme-specific knowledge learned by fine-tuning, we calculate the difference between the fine-tuned CLIP embedding of an image and the embedding from the base CLIP model.

\paragraph{CLIP-diff + RoBERTa.} Finally, we concatenate the CLIP-diff and RoBERTa embeddings to incorporate both the visual features from CLIP as well as the semantics learned by the RoBERTa model.

\subsubsection{Semantic clusters}\label{variables}

To group templates into meme variables, we use Leiden clustering on the template representations from the embedding models. Following literature on using embeddings in later layers of language models \cite{timkey-van-schijndel-2021-bark}, we first standardized the template embeddings before calculating the cosine similarity.

We construct an adjacency matrix from the top 50 nearest neighbors for each template embedding, weighting edges as a function of the ranked cosine similarity. We provide details about this process in Appendix~\ref{sec:sem_details}. We generate \textbf{semantic clusters} by running the Leiden algorithm on this graph.

\section{\textsc{SemanticMemes} Dataset}

We used the pipeline described above to generate semantic clusters from a dataset of 27.9M images collected from Reddit. We scraped images from the top 1000 most active subreddits with ``meme'' in name (e.g. \texttt{r/HistoryMemes}). Temporally, the dataset spans the decade between 2011 and mid-2021. We save metadata from each post including the author, timestamp, and subreddit.

We fine-tuned both the RoBERTa and CLIP models for three epochs on memes whose template appeared at least 100 times in the dataset. We used an 80/10/10 split of train, dev and test data, ensuring there was no leakage of fill text between splits.

Using the pipeline with the RoBERTa model results in 784 semantic clusters spanning 6,384 templates and over 3.8M meme instances. Figure~\ref{fig:cluster-examples} shows some templates that appear in the same semantic cluster. The dataset includes posts to 655 subreddits by 908,917 users. We include examples and descriptive statistics for clusters generated with each of the embedding models in the appendix.

\section{Evaluation}

We evaluate the coherence and visual diversity of semantic clusters derived from each model using human judgment. We design an evaluation task in which annotators are presented with a pair of templates, and randomly vary if the templates are drawn from the same or different semantic clusters.

They are asked to evaluate whether the two templates are 1) semantically similar and 2) visually similar. We define semantic similarity as being able to reasonably substitute the text from one template into the other with minor changes. We define visual similarity to include sharing a similar art style or source (e.g., two different templates featuring Spongebob). We include example pairs in the appendix; one strong source of visual similarity (cf. Appendix Fig.~\ref{fig:clip_sem}) are sets of templates that are largely identical in their form but that exhibit slight variation in size, crop, and margins.

We collect judgments for the top ten semantic clusters from each model most commonly represented in our dataset as well as a random selection of ten clusters from each model. For each cluster, we sample 10 pairs. We find strong interannotator agreement (Krippendorff's $\alpha$=0.75).  From the human judgments, we calculate $p_s$ (the probability that a pair of templates are semantically similar if they appear in the same cluster) and $p_v$ (the equivalent measurement for visual similarity) for each model. We use precision as the evaluation metric because, to measure variation, it is more important each semantic cluster is semantically coherent and visually diverse, but less important that all relevant templates are surfaced within the same cluster.%

Our goal in this work is to explore meaningful semantic variation across visually \emph{diverse} memes, since memes that are visually similar (e.g., slight variations on the same template) have trivially similar semantics. Accordingly, we design a measure of
\textit{visually adjusted precision} 
based on Cohen's $\kappa$: \[
    p_{\text{adj}} = \frac{p_s - p_v}{1 - p_v}
,\]

Intuitively, this metric represents the extent to which the semantic clusters agree with annotator judgments of semantic similarity while controlling for correlations with visual similarity. A negative score means the model clusters based on visual similarity instead of semantic coherence.%

\begin{table}
    \centering
    \begin{tabular}{lrr}
        \textbf{Model} & \textbf{Precision} & \textbf{Visual-adjusted} \\
        \hline
        {RoBERTa}    & \textbf{0.78} & \textbf{0.44}\\
        {CLIP}       & {0.65} & -0.09\\
        {CLIP-diff}  & 0.69 & 0.18 \\
        {Concat.}    & 0.70 & 0.30 \\
    \end{tabular}
    \caption{Comparison of cluster quality for different embedding models. CLIP-based models yield clusters that are biased towards visual features.}
    \label{tab:perf}
\end{table}

Table~\ref{tab:perf}  presents the results of this evaluation. We find that introducing any visual features result in some clusters based on visual similarity instead of semantic similarity; accordingly, clusters generated from RoBERTa embeddings have the highest visually-adjusted precision. 

Semantic clusters provide a strong separation between content (the semantic cluster) and style (the choice of template within a semantic cluster). In other words, the choice of semantic cluster is \textit{what} a user is trying to say, and the choice of a template within that cluster is \textit{how} they are saying it. In the remainder of the paper, we use the clusters generated from the RoBERTa embeddings, which have the highest visual-adjusted precision, for our case studies on linguistic variation and change.

\section{Linguistic variation}
\label{sec:variation}

The sociolinguistic study of variation centers around the linguistic variable, which captures different ways of saying the same thing. The specific choice a speaker make varies systematically based on information such as the speaker's identity, their relationship to interlocutors, sociopragmatic context, among many other factors \cite{tagliamonteAnalysingSociolinguisticVariation2006}. Through variation, language conveys \textit{social meaning} \cite{nguyenLearningRepresentingSocial2021}.

There is a rich body of work that aims to analyze linguistic variation computationally. 
Often, the focus is on lexical variation \cite{bammanGenderIdentityLexical2014a,zhangCommunityIdentityUser2017a,zhu2021idiosyncratic};  semantic variation in online communities \cite{lucy-bamman-2021-characterizing, deltrediciSemanticVariationOnline2018}; or orthographic variation in online text \cite{eisensteinSystematicPatterningPhonologicallymotivated2015, stewartAnorexiaAnarexiaAnarexyia2017}.  In our view of memes as language, we ask the same kind of question:

\paragraph{RQ1:} Does the template choice within a semantic cluster vary systematically between communities?

\begin{figure*}[ht]
        \centering
        \fbox*[boxsep=0.0\linewidth,boxrule=0pt]{\begin{subfigure}[b]{\dimexpr 0.4\linewidth}
            \centering
            \begin{subfigure}{1.0in}
                \caption*{\texttt{r/memes}}            
                \fbox*[boxsep=0.0pt]{\includegraphics[width=\linewidth]{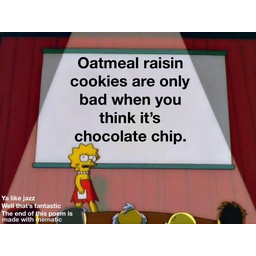}}
            \end{subfigure}%
            \begin{subfigure}{1.0in}
                \caption*{\texttt{r/Animemes}}            
                \fbox*[boxsep=0.0pt]{\includegraphics[width=\linewidth]{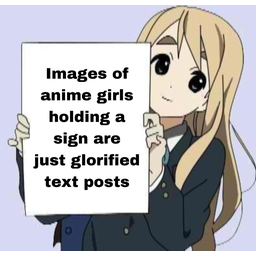}}
            \end{subfigure}
            \caption{Declarative}
        \end{subfigure}}%
        \fbox*[boxsep=0.0\linewidth,boxrule=0pt]{\begin{subfigure}[b]{\dimexpr 0.6\linewidth}
            \centering
            \begin{subfigure}{1.0in}
                \caption*{\texttt{r/memes}}            
                \fbox*[boxsep=0.0pt]{\includegraphics[width=\linewidth]{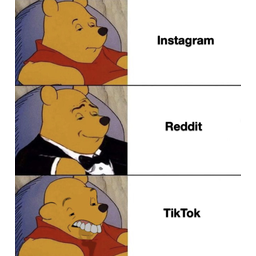}}
            \end{subfigure}%
            \begin{subfigure}{1.0in}
                \caption*{\texttt{r/dndmemes}}            
                \fbox*[boxsep=0.0pt]{\includegraphics[width=\linewidth]{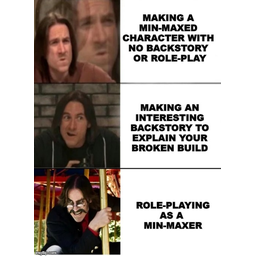}}
            \end{subfigure}%
            \begin{subfigure}{1.0in}
                \caption*{\texttt{r/MinecraftMemes}}            
                \fbox*[boxsep=0.0pt]{\includegraphics[width=\linewidth]{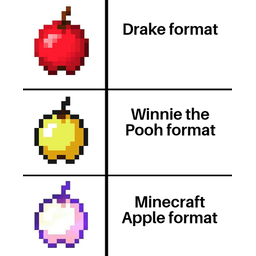}}
            \end{subfigure}
            \caption{Scalar increase}
        \end{subfigure}}
        \vspace{2pt}
        \begin{subfigure}[b]{\dimexpr \linewidth}
            \centering
            
            \begin{subfigure}{1.0in}
                \caption*{\texttt{r/memes}}            
                \fbox*[boxsep=0.0pt]{\includegraphics[width=\linewidth]{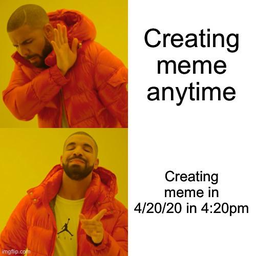}}
            \end{subfigure}%
            \begin{subfigure}{1.0in}
                \caption*{\texttt{r/prequelmemes}}            
                \fbox*[boxsep=0.0pt]{\includegraphics[width=\linewidth]{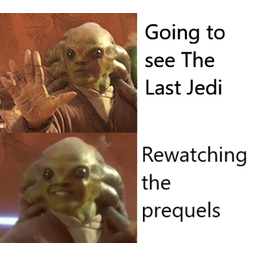}}
            \end{subfigure}%
            \begin{subfigure}{1.0in}
                \caption*{\texttt{r/dndmemes}}            
                \fbox*[boxsep=0.0pt]{\includegraphics[width=\linewidth]{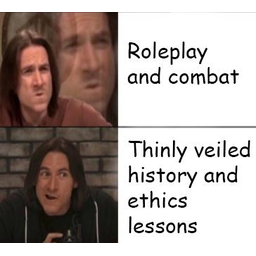}}
            \end{subfigure}%
            \begin{subfigure}{1.0in}
                \caption*{\texttt{r/Animemes}}            
                \fbox*[boxsep=0.0pt]{\includegraphics[width=\linewidth]{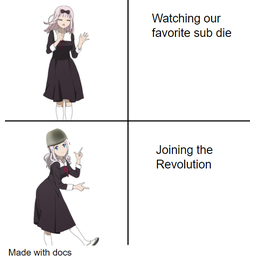}}
            \end{subfigure}%
            \begin{subfigure}{1.0in}
                \caption*{\texttt{r/startrekmemes}}            
                \fbox*[boxsep=0.0pt]{\includegraphics[width=\linewidth]{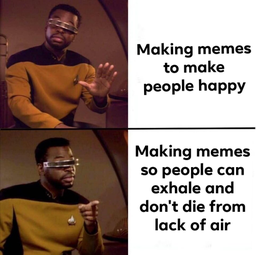}}
            \end{subfigure}%
            \caption{Comparison}
            \label{fig:variation_comparison}
        \end{subfigure}%

    \caption{Subreddits exhibit variation in the preferred templates within a semantic cluster. All are statistically significantly overrepresented in their respective subreddits, $p < 0.05$.}
    \label{fig:variation}
\end{figure*}

\paragraph{Methods.}
The semantic clusters form our variable context, and set of templates within any given semantic cluster form a discrete set of choices with the same semantic value.
We use the weighted log odds-ratio to compute the extent to which a template is specific to a given subreddit compared to all other subreddits, relative to the other templates in a semantic cluster \cite{monroeFightinWordsLexical2017, jurafskyNarrativeFramingConsumer2014}. We find the templates that have a statistically significant association with a subreddit ($z$-score $> 1.96$); the semantic clusters these templates belong to are \textit{in variation}: a community prefers one variant over the others in this cluster.

\paragraph{Results.}

We find 94 out of 784 semantic clusters exhibit statistically significant variation, spanning 391 different templates.
Figure~\ref{fig:variation} shows how functionally similar memes take different forms in different communities.

Speakers use language to construct their social identities \cite{bucholtzIdentityInteractionSociocultural2005}.
We find that, not only do subreddits prefer certain variants of a template over others, but they choose templates that index into a localized cultural knowledge, making cultural allusions to characters or celebrities. %

For example, the orange Drake template (fig.~\ref{fig:variation_comparison}, left) is used frequently in general purpose meme subreddits like \texttt{r/memes}, but alternatives are used in other subreddits. One version that is specific to \texttt{r/dndmemes} (which discusses the role playing game Dungeons and Dragons) replaces Drake with Matthew Mercer, a voice actor who stars in a popular Dungeons and Dragons web series (fig.~\ref{fig:variation_comparison}, middle).

Linguistic variants usually become associated with identities through a gradual process in which the association slowly permeates public awareness \cite{eckertVariationIndexicalField2008}. In general, a phonological variable does not inherently index any given identity. However, the multimodality of memes permits greater expressiveness---a meme in \texttt{r/Animemes} might use the anime art style, indexing into the aesthetic of that community explicitly.

\section{Linguistic innovation}
Equally as important as the study of synchronic linguistic variation is the study of diachronic linguistic change. Language change has been heavily studied in natural language processing
\cite{rosenfeld-erk-2018-deep, martinc-etal-2020-leveraging,zhu2021structure}. We focus on understanding the innovation of meme templates within a semantic cluster.

\paragraph{RQ2:} Do new meme templates co-exist with preexisting templates in the semantic cluster, or does the most popular template monopolize the cluster?

When multiple templates that fulfill the same function appear, we expect there to be competition. Prior work has observed this competition between lexical choices, with two outcomes: new words replace old ones that serve the same function, but if similar words have discourse-relevant differences in meaning, they can coexist \cite{karjusCommunicativeNeedModulates2020}.

\paragraph{Methods}

We measure the entropy of semantic clusters over time. If meme templates ultimately co-exist, we would expect entropy to increase; if a subset of templates dominate, we would expect the entropy to converge to a lower value. %

For each semantic cluster, we group posts by the age of the semantic cluster in years at the time of posting. We define the ``birth'' of the cluster as when the first instance of a template in that cluster was posted. Within each year, we calculate the entropy of template distribution within each cluster.

It is possible that some clusters have low entropy early on due to data sparsity. To account for this, we filter to semantic clusters that have existed at least 5 years with at least 30 posts in all years, and resample with replacement within each year such that every year has the same number of posts. Ultimately, we conduct our analysis over 146 semantic clusters that span over 950K posts.

\paragraph{Results}

\begin{figure}
    \centering
    \includegraphics[width=\linewidth]{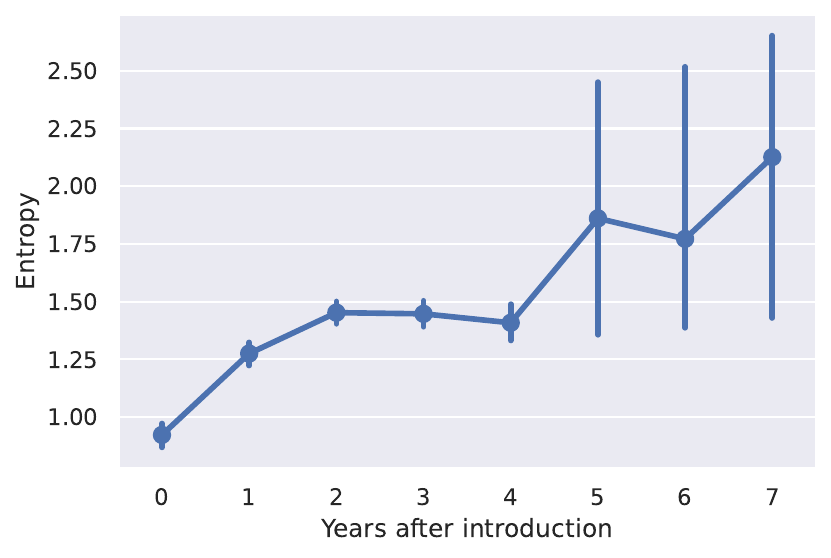}
    \caption{On average, semantic clusters diversify over time. Very old semantic clusters are rarer, leading to larger confidence intervals in later years.}
    \label{fig:entropy}
\end{figure}

Entropy steadily increases in the years following a semantic cluster's initial introduction (Figure~\ref{fig:entropy}). This suggests that no one meme template grows to become the de facto template for all users; there is steady variation. This is supported by our findings in Section~\ref{sec:variation} that there are socially meaningful differences between variants.

\paragraph{RQ3:} Do new templates diffuse widely or occupy a niche?

Language change is often socially motivated; a community can opt to use a particular variant to distinguish themselves from others \cite{trudgillDialectsContact1986, gilesAccommodationTheory1975}. Thus, we might expect meme templates to be most specific to the subreddits in which they were first introduced.

\paragraph{Methods}

We measure the extent to which template variants are ultimately specific to the subreddits that originated them.

We filter our dataset to templates which occur at least 200 times. For each template, we identify a set of ``seed posts'', which we define as the first 100 posts using the template. We then filter to templates with a subreddit that comprises the majority of the seed posts, which we call the ``origin subreddit.''

We modify the method from \cite{zhangCommunityIdentityUser2017a} to measure the specificity of a template-subreddit pair by using the {positive} pointwise mutual information (PPMI) between templates and the subreddits in which they are used, matching other work in NLP \cite{church-hanks-1990-word, slp3}. Formally, we calculate
\[
  \text{PPMI}(t;s \mid c) = \max\left(\log {\frac{P(t \mid s, c)}{P(t \mid c)}}, 0\right)
,\]
where $P(t \mid s, c)$ is the probability of template $t$ appearing in subreddit $s$ and semantic cluster $c$, $P(t \mid c)$ is the probability of template $t$ in that cluster globally, and templates are only compared against others within the same semantic cluster. We calculate the PPMI over non-seed posts to measure the specificity of a template after its introduction. %

\paragraph{Results}

\begin{figure}
    \centering
    \includegraphics[width=1\linewidth]{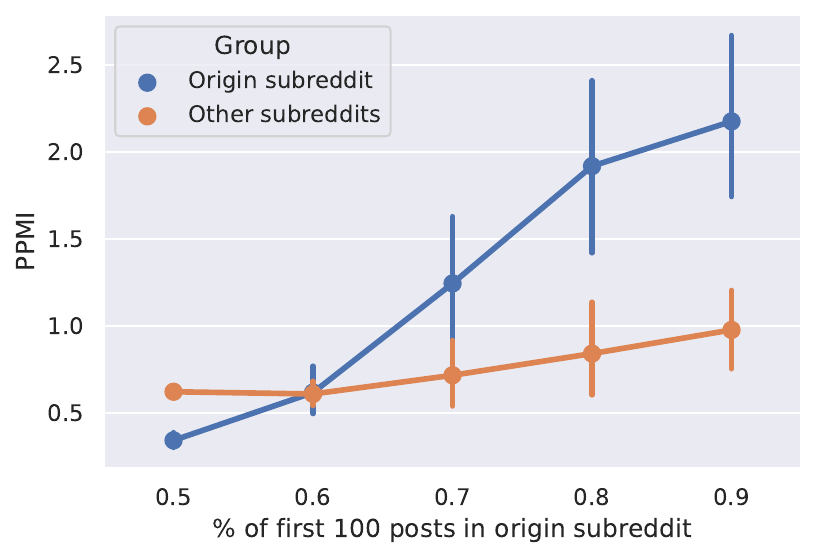}
    \caption{Communities that lead the introduction of a new template continue to use it more than others.}
    \label{fig:divergence}
\end{figure}

For each template, we compare the PPMI for origin subreddits with the average PPMI of all other subreddits. Figure~\ref{fig:divergence} shows a significant positive correlation between the proportion of seed posts that originated in the origin subreddit and the eventual specificity of template. These results support previous findings that lexical innovations succeed when filling in a social niche \cite{altmannNicheDeterminantWord2011, macwhinneyCompetitionLexicalCategorization1989}.

\section{Linguistic acculturation}

Finally, we study how users alter their meme posting habits as they spend more time in a subreddit. Previous work on linguistic acculturation show that users adopt more community-specific language as they become enculturated within a community \cite{danescu-niculescu-mizilNoCountryOld2013, srivastavaEnculturationTrajectoriesLanguage2018}. We can ask a  similar question here:

\paragraph{RQ4:} Do veteran users in a subreddit use more community-specific templates?

\paragraph{Methods}

To answer this question, we measure the average specificity of a user's posts in successive months after they enter a community. We once again calculate the PPMI of templates as a measure of specificity; this time, we calculate the value over the full range of the dataset.

For each user in a subreddit, we bin their posts by 30-day windows starting with their first post in the subreddit (i.e., for each month after they joined), and compute the average PPMI of their posts for that time period. We filter the dataset to users with at least 10 lifetime posts and subreddits with at least 30 such users. To prevent extremely popular subreddits from unduly influencing the results, we sample up to 100 users from each subreddit to compute the average across all subreddits. This yields a total of 3,174 users in 130 subreddits.

\paragraph{Results}

\begin{figure}
    \centering
    \includegraphics[width=1\linewidth]{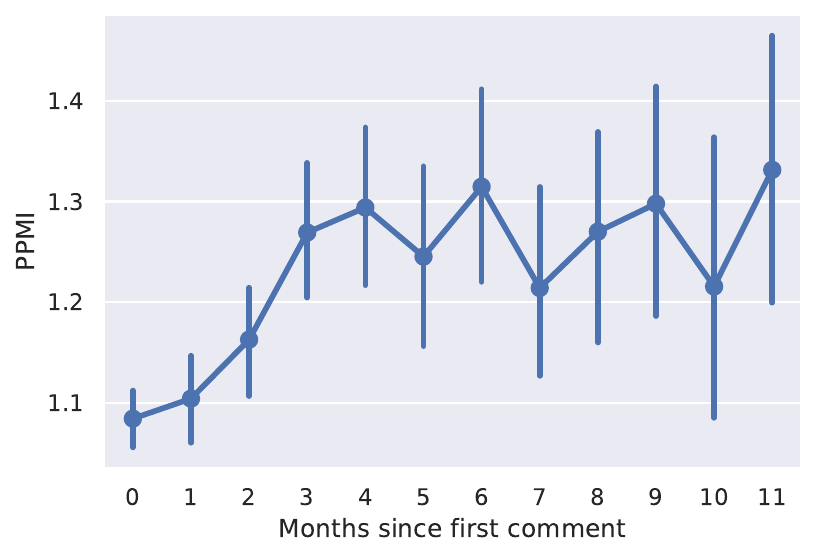}
    \caption{Veteran (acculturated) users employ more subreddit-specific meme templates.}
    \label{fig:acculturation}
\end{figure}

We find that acculturated users use templates that are more specific to the communities in which they  post (Pearson's $r = 0.074, p < 0.001$), shown in Figure \ref{fig:acculturation}.
Our finding aligns with existing literature on linguistic acculturation as well as theories in new media that memes are cultural capital. The ``correct'' use of memes can demonstrate a user's assimilation into a shared language and identity \cite{nissenbaumInternetMemesContested2017}.%

\section{Related work}

Prior work on memes in NLP and social computing has largely focused on two  tasks: meme understanding and modeling how memes originate and spread. Our work offers novel methods and perspectives at the intersection of these  areas of research.

Meme understanding encompasses a number of discrete tasks, including classifying if memes convey harmful messages \cite{kielaHatefulMemesChallenge2021, quDisinfoMemeMultimodalDataset2022},  labeling emotion \cite{mishraMemotionDatasetSentiment2023}, and detecting humor \cite{tanaka-etal-2022-learning} or figurative speech \cite{liu-etal-2022-figmemes} within them.  While these can generally be framed as classification tasks, other work generates open-ended explanations of visual humor using large multimodal language models \cite{hwangMemeCapDatasetCaptioning2023, hessel-etal-2023-androids}. Our work complements this existing body of research by inferring semantic variables in an unsupervised approach, leveraging the implicit structure within memes by modeling modeling template semantics separately from the fills.

In modeling the internal structure of memes, our work draws on existing research examining the relationship between fills and templates to match semantic roles to entities within harmful memes \cite{sharma-etal-2023-characterizing} and mapping fill text to explanatory background information \cite{sharma-etal-2023-memex}. We hope that our method of construing templates as semantic predicates can contribute to this body of work.

In the social computing space, another line of research focuses on understanding how memes originate \cite{Morina2022} and spread across platforms \cite{Zannettou2018}. These treat meme templates as discrete tokens. We model template semantics, which have the granularity to enable analysis of variation and social meaning. \citeauthor{quEvolutionHatefulMemes2023} uses CLIP to understand how memes evolve as they spread. While they use the text in comments to model the high-level concepts indexed by particular variants, we use the fill text of memes to model low-level template semantics.

\section{Conclusion}

In this paper, we analyze memes as a form of language subject to the same kinds of sociolinguistic variation as other modalities, such as written language and speech.
We propose a new approach to understanding meme semantics, taking advantage of the multimodal structure of memes to learn semantic representations of templates from an unlabeled dataset. We use this method on a large dataset of memes scraped from Reddit, and demonstrate that it yields coherent, visually diverse clusters of semantically similar memes. We make these clusters and the code publicly available for future research. Finally, we use these clusters to study language variation and change in subreddits. We show that variations between meme template are socially meaningful and memes often share usage patterns with the textual language that has been studied in the past. We find that memes can be rich resources for understanding social language use.%

\section{Ethical considerations}

The data used in this work was collected from Reddit in 2021 and is publicly available. There may be offensive, hateful, or sexual messages present in the memes and comments in this dataset.

The models we trained are also publicly available. We use them only to better understand the semantics of memes. We do not train any generative models, and warn against training generative models on the data without careful consideration of how to mitigate the toxic, offensive, or otherwise harmful outputs that might be generated.

\section*{Acknowledgments}

The research reported in this article was supported by funding from the National Science Foundation (Graduate Research Fellowship DGE-2146752 and grant IIS-1942591) and from the Volkswagen Foundation.

\bibliography{anthology,custom}

\begin{thebibliography}{52}
\expandafter\ifx\csname natexlab\endcsname\relax\def\natexlab#1{#1}\fi

\bibitem[{Altmann et~al.(2011)Altmann, Pierrehumbert, and
  Motter}]{altmannNicheDeterminantWord2011}
Eduardo~G. Altmann, Janet~B. Pierrehumbert, and Adilson~E. Motter. 2011.
\newblock \href {https://doi.org/10.1371/journal.pone.0019009} {Niche as a
  {{Determinant}} of {{Word Fate}} in {{Online Groups}}}.
\newblock \emph{PLOS ONE}, 6(5):e19009.

\bibitem[{Bamman et~al.(2014)Bamman, Eisenstein, and
  Schnoebelen}]{bammanGenderIdentityLexical2014a}
David Bamman, Jacob Eisenstein, and Tyler Schnoebelen. 2014.
\newblock \href {https://doi.org/10.1111/josl.12080} {Gender identity and
  lexical variation in social media}.
\newblock \emph{Journal of Sociolinguistics}, 18(2):135--160.

\bibitem[{Blodgett et~al.(2016)Blodgett, Green, and
  O{'}Connor}]{blodgett-etal-2016-demographic}
Su~Lin Blodgett, Lisa Green, and Brendan O{'}Connor. 2016.
\newblock \href {https://doi.org/10.18653/v1/D16-1120} {Demographic dialectal
  variation in social media: A case study of {A}frican-{A}merican {E}nglish}.
\newblock In \emph{Proceedings of the 2016 Conference on Empirical Methods in
  Natural Language Processing}, pages 1119--1130, Austin, Texas. Association
  for Computational Linguistics.

\bibitem[{Bucholtz and
  Hall(2005)}]{bucholtzIdentityInteractionSociocultural2005}
Mary Bucholtz and Kira Hall. 2005.
\newblock \href {https://doi.org/10.1177/1461445605054407} {Identity and
  interaction: A sociocultural linguistic approach}.
\newblock \emph{Discourse Studies}, 7(4-5):585--614.

\bibitem[{{Campbell-Kibler}(2009)}]{campbell-kiblerNatureSociolinguisticPerception2009}
Kathryn {Campbell-Kibler}. 2009.
\newblock \href {https://doi.org/10.1017/S0954394509000052} {The nature of
  sociolinguistic perception}.
\newblock \emph{Language Variation and Change}, 21(1):135--156.

\bibitem[{Church and Hanks(1990)}]{church-hanks-1990-word}
Kenneth~Ward Church and Patrick Hanks. 1990.
\newblock \href {https://aclanthology.org/J90-1003} {Word association norms,
  mutual information, and lexicography}.
\newblock \emph{Computational Linguistics}, 16(1):22--29.

\bibitem[{{Danescu-Niculescu-Mizil} et~al.(2013){Danescu-Niculescu-Mizil},
  West, Jurafsky, Leskovec, and
  Potts}]{danescu-niculescu-mizilNoCountryOld2013}
Cristian {Danescu-Niculescu-Mizil}, Robert West, Dan Jurafsky, Jure Leskovec,
  and Christopher Potts. 2013.
\newblock \href {https://doi.org/10.1145/2488388.2488416} {No country for old
  members: User lifecycle and linguistic change in online communities}.
\newblock In \emph{Proceedings of the 22nd International Conference on {{World
  Wide Web}}}, {{WWW}} '13, pages 307--318, {New York, NY, USA}. {Association
  for Computing Machinery}.

\bibitem[{Del~Tredici and
  Fern{\'a}ndez(2017)}]{del-tredici-fernandez-2017-semantic}
Marco Del~Tredici and Raquel Fern{\'a}ndez. 2017.
\newblock \href {https://aclanthology.org/W17-6804} {Semantic variation in
  online communities of practice}.
\newblock In \emph{{IWCS} 2017 - 12th International Conference on Computational
  Semantics - Long papers}.

\bibitem[{Del~Tredici and
  Fern{\'a}ndez(2018)}]{deltrediciSemanticVariationOnline2018}
Marco Del~Tredici and Raquel Fern{\'a}ndez. 2018.
\newblock \href {http://arxiv.org/abs/1806.05847} {Semantic {{Variation}} in
  {{Online Communities}} of {{Practice}}}.
\newblock \emph{arXiv:1806.05847 [cs]}.

\bibitem[{Demszky et~al.(2021)Demszky, Sharma, Clark, Prabhakaran, and
  Eisenstein}]{demszky-etal-2021-learning}
Dorottya Demszky, Devyani Sharma, Jonathan Clark, Vinodkumar Prabhakaran, and
  Jacob Eisenstein. 2021.
\newblock \href {https://doi.org/10.18653/v1/2021.naacl-main.184} {Learning to
  recognize dialect features}.
\newblock In \emph{Proceedings of the 2021 Conference of the North American
  Chapter of the Association for Computational Linguistics: Human Language
  Technologies}, pages 2315--2338, Online. Association for Computational
  Linguistics.

\bibitem[{Eckert(2008)}]{eckertVariationIndexicalField2008}
Penelope Eckert. 2008.
\newblock \href {https://doi.org/10.1111/j.1467-9841.2008.00374.x} {Variation
  and the indexical field}.
\newblock \emph{Journal of Sociolinguistics}, 12(4):453--476.

\bibitem[{Eisenstein(2015)}]{eisensteinSystematicPatterningPhonologicallymotivated2015}
Jacob Eisenstein. 2015.
\newblock \href {https://doi.org/10.1111/josl.12119} {Systematic patterning in
  phonologically-motivated orthographic variation}.
\newblock \emph{Journal of Sociolinguistics}, 19(2):161--188.

\bibitem[{Eisenstein et~al.(2010)Eisenstein, O{'}Connor, Smith, and
  Xing}]{eisenstein-etal-2010-latent}
Jacob Eisenstein, Brendan O{'}Connor, Noah~A. Smith, and Eric~P. Xing. 2010.
\newblock \href {https://aclanthology.org/D10-1124} {A latent variable model
  for geographic lexical variation}.
\newblock In \emph{Proceedings of the 2010 Conference on Empirical Methods in
  Natural Language Processing}, pages 1277--1287, Cambridge, MA. Association
  for Computational Linguistics.

\bibitem[{Giles and Powesland(1975)}]{gilesAccommodationTheory1975}
Howard Giles and Peter~F. Powesland. 1975.
\newblock Accommodation {{Theory}}.
\newblock In \emph{Speech Style and Social Evaluation}, Speech Style and Social
  Evaluation, pages 232--239. {Academic Press}, {Oxford, England}.

\bibitem[{Hamilton et~al.(2016)Hamilton, Leskovec, and
  Jurafsky}]{hamilton-etal-2016-diachronic}
William~L. Hamilton, Jure Leskovec, and Dan Jurafsky. 2016.
\newblock \href {https://doi.org/10.18653/v1/P16-1141} {Diachronic word
  embeddings reveal statistical laws of semantic change}.
\newblock In \emph{Proceedings of the 54th Annual Meeting of the Association
  for Computational Linguistics (Volume 1: Long Papers)}, pages 1489--1501,
  Berlin, Germany. Association for Computational Linguistics.

\bibitem[{Hessel et~al.(2023)Hessel, Marasovic, Hwang, Lee, Da, Zellers,
  Mankoff, and Choi}]{hessel-etal-2023-androids}
Jack Hessel, Ana Marasovic, Jena~D. Hwang, Lillian Lee, Jeff Da, Rowan Zellers,
  Robert Mankoff, and Yejin Choi. 2023.
\newblock \href {https://aclanthology.org/2023.acl-long.41} {Do androids laugh
  at electric sheep? humor {``}understanding{''} benchmarks from the new yorker
  caption contest}.
\newblock In \emph{Proceedings of the 61st Annual Meeting of the Association
  for Computational Linguistics (Volume 1: Long Papers)}, pages 688--714,
  Toronto, Canada. Association for Computational Linguistics.

\bibitem[{Hovy and Purschke(2018)}]{hovy-purschke-2018-capturing}
Dirk Hovy and Christoph Purschke. 2018.
\newblock \href {https://doi.org/10.18653/v1/D18-1469} {Capturing regional
  variation with distributed place representations and geographic
  retrofitting}.
\newblock In \emph{Proceedings of the 2018 Conference on Empirical Methods in
  Natural Language Processing}, pages 4383--4394, Brussels, Belgium.
  Association for Computational Linguistics.

\bibitem[{Hwang and Shwartz(2023)}]{hwangMemeCapDatasetCaptioning2023}
EunJeong Hwang and Vered Shwartz. 2023.
\newblock \href {https://doi.org/10.48550/arXiv.2305.13703} {{MemeCap}: {A}
  {Dataset} for {Captioning} and {Interpreting} {Memes}}.
\newblock ArXiv:2305.13703 [cs].

\bibitem[{Jurafsky et~al.(2014)Jurafsky, Chahuneau, Routledge, and
  Smith}]{jurafskyNarrativeFramingConsumer2014}
Dan Jurafsky, Victor Chahuneau, Bryan~R. Routledge, and Noah~A. Smith. 2014.
\newblock \href {https://doi.org/10.5210/fm.v19i4.4944} {Narrative framing of
  consumer sentiment in online restaurant reviews}.
\newblock \emph{First Monday}.

\bibitem[{Jurafsky and Martin(2009)}]{slp3}
Daniel Jurafsky and James~H. Martin. 2009.
\newblock \emph{Speech and Language Processing (2nd Edition)}.
\newblock Prentice-Hall, Inc., USA.

\bibitem[{Karjus et~al.(2020)Karjus, Blythe, Kirby, and
  Smith}]{karjusCommunicativeNeedModulates2020}
Andres Karjus, Richard~A. Blythe, Simon Kirby, and Kenny Smith. 2020.
\newblock \href {http://arxiv.org/abs/2006.09277} {Communicative need modulates
  competition in language change}.

\bibitem[{Kiela et~al.(2021)Kiela, Firooz, Mohan, Goswami, Singh, Ringshia, and
  Testuggine}]{kielaHatefulMemesChallenge2021}
Douwe Kiela, Hamed Firooz, Aravind Mohan, Vedanuj Goswami, Amanpreet Singh,
  Pratik Ringshia, and Davide Testuggine. 2021.
\newblock \href {https://doi.org/10.48550/arXiv.2005.04790} {The {Hateful}
  {Memes} {Challenge}: {Detecting} {Hate} {Speech} in {Multimodal} {Memes}}.
\newblock ArXiv:2005.04790 [cs].

\bibitem[{Kress and Leeuwen(2001)}]{kress2001}
Gunther~R. Kress and Theo~Van Leeuwen. 2001.
\newblock \emph{Multimodal Discourse: The Modes and Media of Contemporary
  Communication}.
\newblock Arnold; Oxford University Press.

\bibitem[{Labov(1963)}]{labovSocialMotivationSound1963a}
William Labov. 1963.
\newblock \href {https://doi.org/10.1080/00437956.1963.11659799} {The {{Social
  Motivation}} of a {{Sound Change}}}.
\newblock \emph{\emph{WORD}}, 19(3):273--309.

\bibitem[{Liu et~al.(2022)Liu, Geigle, Krebs, and
  Gurevych}]{liu-etal-2022-figmemes}
Chen Liu, Gregor Geigle, Robin Krebs, and Iryna Gurevych. 2022.
\newblock \href {https://aclanthology.org/2022.emnlp-main.476} {{F}ig{M}emes: A
  dataset for figurative language identification in politically-opinionated
  memes}.
\newblock In \emph{Proceedings of the 2022 Conference on Empirical Methods in
  Natural Language Processing}, pages 7069--7086, Abu Dhabi, United Arab
  Emirates. Association for Computational Linguistics.

\bibitem[{Lucy and Bamman(2021)}]{lucy-bamman-2021-characterizing}
Li~Lucy and David Bamman. 2021.
\newblock \href {https://doi.org/10.1162/tacl_a_00383} {Characterizing
  {E}nglish variation across social media communities with {BERT}}.
\newblock \emph{Transactions of the Association for Computational Linguistics},
  9:538--556.

\bibitem[{MacWhinney(1989)}]{macwhinneyCompetitionLexicalCategorization1989}
Brian MacWhinney. 1989.
\newblock \href {https://doi.org/10.1075/cilt.61.14mac} {Competition and
  lexical categorization}.
\newblock In Roberta Corrigan, Fred~R. Eckman, and Michael Noonan, editors,
  \emph{Current {{Issues}} in {{Linguistic Theory}}}, volume~61, page 195.
  {John Benjamins Publishing Company}, {Amsterdam}.

\bibitem[{Martinc et~al.(2020)Martinc, Kralj~Novak, and
  Pollak}]{martinc-etal-2020-leveraging}
Matej Martinc, Petra Kralj~Novak, and Senja Pollak. 2020.
\newblock \href {https://aclanthology.org/2020.lrec-1.592} {Leveraging
  contextual embeddings for detecting diachronic semantic shift}.
\newblock In \emph{Proceedings of the Twelfth Language Resources and Evaluation
  Conference}, pages 4811--4819, Marseille, France. European Language Resources
  Association.

\bibitem[{Mishra et~al.(2023)Mishra, Suryavardan, Patwa, Chakraborty, Rani,
  Reganti, Chadha, Das, Sheth, Chinnakotla, Ekbal, and
  Kumar}]{mishraMemotionDatasetSentiment2023}
Shreyash Mishra, S.~Suryavardan, Parth Patwa, Megha Chakraborty, Anku Rani,
  Aishwarya Reganti, Aman Chadha, Amitava Das, Amit Sheth, Manoj Chinnakotla,
  Asif Ekbal, and Srijan Kumar. 2023.
\newblock \href {https://doi.org/10.48550/arXiv.2303.09892} {Memotion 3:
  {Dataset} on {Sentiment} and {Emotion} {Analysis} of {Codemixed}
  {Hindi}-{English} {Memes}}.
\newblock ArXiv:2303.09892 [cs].

\bibitem[{Monroe et~al.(2017)Monroe, Colaresi, and
  Quinn}]{monroeFightinWordsLexical2017}
Burt~L. Monroe, Michael~P. Colaresi, and Kevin~M. Quinn. 2017.
\newblock \href {https://doi.org/10.1093/pan/mpn018} {Fightin' {{Words}}:
  {{Lexical Feature Selection}} and {{Evaluation}} for {{Identifying}} the
  {{Content}} of {{Political Conflict}}}.
\newblock \emph{Political Analysis}, 16(4):372--403.

\bibitem[{Morina and Bernstein(2022)}]{Morina2022}
Durim Morina and Michael~S. Bernstein. 2022.
\newblock \href {https://doi.org/10.1145/3512921} {A web-scale analysis of the
  community origins of image memes}.
\newblock \emph{Proc. ACM Hum.-Comput. Interact.}, 6(CSCW1).

\bibitem[{Nguyen et~al.(2021)Nguyen, Rosseel, and
  Grieve}]{nguyenLearningRepresentingSocial2021}
Dong Nguyen, Laura Rosseel, and Jack Grieve. 2021.
\newblock \href {https://doi.org/10.18653/v1/2021.naacl-main.50} {On learning
  and representing social meaning in {NLP}: a sociolinguistic perspective}.
\newblock In \emph{Proceedings of the 2021 {Conference} of the {North}
  {American} {Chapter} of the {Association} for {Computational} {Linguistics}:
  {Human} {Language} {Technologies}}, pages 603--612, Online. Association for
  Computational Linguistics.

\bibitem[{Nissenbaum and Shifman(2017)}]{nissenbaumInternetMemesContested2017}
Asaf Nissenbaum and Limor Shifman. 2017.
\newblock \href {https://doi.org/10.1177/1461444815609313} {Internet memes as
  contested cultural capital: {{The}} case of 4chan's /b/ board}.
\newblock \emph{New Media \& Society}, 19(4):483--501.

\bibitem[{Perniss(2018)}]{pernissWhyWeShould2018}
Pamela Perniss. 2018.
\newblock Why {{We Should Study Multimodal Language}}.
\newblock \emph{Frontiers in Psychology}, 9.

\bibitem[{Qu et~al.(2022)Qu, Li, Zhao, Dev, and
  Chang}]{quDisinfoMemeMultimodalDataset2022}
Jingnong Qu, Liunian~Harold Li, Jieyu Zhao, Sunipa Dev, and Kai-Wei Chang.
  2022.
\newblock \href {https://doi.org/10.48550/arXiv.2205.12617} {{{DisinfoMeme}}:
  {{A Multimodal Dataset}} for {{Detecting Meme Intentionally Spreading Out
  Disinformation}}}.

\bibitem[{Qu et~al.(2023)Qu, He, Pierson, Backes, Zhang, and
  Zannettou}]{quEvolutionHatefulMemes2023}
Yiting Qu, Xinlei He, Shannon Pierson, Michael Backes, Yang Zhang, and Savvas
  Zannettou. 2023.
\newblock \href {https://doi.org/10.1109/SP46215.2023.10179315} {On the
  {{Evolution}} of ({{Hateful}}) {{Memes}} by {{Means}} of {{Multimodal
  Contrastive Learning}}}.
\newblock In \emph{2023 {{IEEE Symposium}} on {{Security}} and {{Privacy}}
  ({{SP}})}, pages 293--310. {IEEE Computer Society}.

\bibitem[{Rosenfeld and Erk(2018)}]{rosenfeld-erk-2018-deep}
Alex Rosenfeld and Katrin Erk. 2018.
\newblock \href {https://doi.org/10.18653/v1/N18-1044} {Deep neural models of
  semantic shift}.
\newblock In \emph{Proceedings of the 2018 Conference of the North {A}merican
  Chapter of the Association for Computational Linguistics: Human Language
  Technologies, Volume 1 (Long Papers)}, pages 474--484, New Orleans,
  Louisiana. Association for Computational Linguistics.

\bibitem[{Sharma et~al.(2023{\natexlab{a}})Sharma, Kulkarni, Suresh, Mathur,
  Nakov, Akhtar, and Chakraborty}]{sharma-etal-2023-characterizing}
Shivam Sharma, Atharva Kulkarni, Tharun Suresh, Himanshi Mathur, Preslav Nakov,
  Md.~Shad Akhtar, and Tanmoy Chakraborty. 2023{\natexlab{a}}.
\newblock \href {https://aclanthology.org/2023.eacl-main.157} {Characterizing
  the entities in harmful memes: Who is the hero, the villain, the victim?}
\newblock In \emph{Proceedings of the 17th Conference of the European Chapter
  of the Association for Computational Linguistics}, pages 2149--2163,
  Dubrovnik, Croatia. Association for Computational Linguistics.

\bibitem[{Sharma et~al.(2023{\natexlab{b}})Sharma, S, Arora, Akhtar, and
  Chakraborty}]{sharma-etal-2023-memex}
Shivam Sharma, Ramaneswaran S, Udit Arora, Md.~Shad Akhtar, and Tanmoy
  Chakraborty. 2023{\natexlab{b}}.
\newblock \href {https://aclanthology.org/2023.acl-long.289} {{MEMEX}:
  Detecting explanatory evidence for memes via knowledge-enriched
  contextualization}.
\newblock In \emph{Proceedings of the 61st Annual Meeting of the Association
  for Computational Linguistics (Volume 1: Long Papers)}, pages 5272--5290,
  Toronto, Canada. Association for Computational Linguistics.

\bibitem[{Srivastava et~al.(2018)Srivastava, Goldberg, Manian, and
  Potts}]{srivastavaEnculturationTrajectoriesLanguage2018}
Sameer~B. Srivastava, Amir Goldberg, V.~Govind Manian, and Christopher Potts.
  2018.
\newblock \href {https://doi.org/10.1287/mnsc.2016.2671} {Enculturation
  {{Trajectories}}: {{Language}}, {{Cultural Adaptation}}, and {{Individual
  Outcomes}} in {{Organizations}}}.
\newblock \emph{Management Science}, 64(3):1348--1364.

\bibitem[{Stewart et~al.(2017)Stewart, Chancellor, De~Choudhury, and
  Eisenstein}]{stewartAnorexiaAnarexiaAnarexyia2017}
Ian Stewart, Stevie Chancellor, Munmun De~Choudhury, and Jacob Eisenstein.
  2017.
\newblock \href {https://doi.org/10.1109/BigData.2017.8258465} {\#{{Anorexia}},
  \#anarexia, \#anarexyia: {{Characterizing}} online community practices with
  orthographic variation}.
\newblock In \emph{2017 {{IEEE International Conference}} on {{Big Data}}
  ({{Big Data}})}, pages 4353--4361.

\bibitem[{Tagliamonte(2006)}]{tagliamonteAnalysingSociolinguisticVariation2006}
S.A. Tagliamonte. 2006.
\newblock \href {https://books.google.com/books?id=3\_uZWdoBsUkC}
  {\emph{Analysing {Sociolinguistic} {Variation}}}.
\newblock Key {Topics} in {Sociolinguistics}. Cambridge University Press.

\bibitem[{Tanaka et~al.(2022)Tanaka, Yamane, Mori, Mukuta, and
  Harada}]{tanaka-etal-2022-learning}
Kohtaro Tanaka, Hiroaki Yamane, Yusuke Mori, Yusuke Mukuta, and Tatsuya Harada.
  2022.
\newblock \href {https://aclanthology.org/2022.cai-1.9} {Learning to evaluate
  humor in memes based on the incongruity theory}.
\newblock In \emph{Proceedings of the Second Workshop on When Creative AI Meets
  Conversational AI}, pages 81--93, Gyeongju, Republic of Korea. Association
  for Computational Linguistics.

\bibitem[{Timkey and van Schijndel(2021)}]{timkey-van-schijndel-2021-bark}
William Timkey and Marten van Schijndel. 2021.
\newblock \href {https://doi.org/10.18653/v1/2021.emnlp-main.372} {All bark and
  no bite: Rogue dimensions in transformer language models obscure
  representational quality}.
\newblock In \emph{Proceedings of the 2021 Conference on Empirical Methods in
  Natural Language Processing}, pages 4527--4546, Online and Punta Cana,
  Dominican Republic. Association for Computational Linguistics.

\bibitem[{Traag et~al.(2019)Traag, Waltman, and van
  Eck}]{traagLouvainLeidenGuaranteeing2019}
V.~A. Traag, L.~Waltman, and N.~J. van Eck. 2019.
\newblock \href {https://doi.org/10.1038/s41598-019-41695-z} {From {Louvain} to
  {Leiden}: guaranteeing well-connected communities}.
\newblock \emph{Scientific Reports}, 9(1):5233.
\newblock Number: 1 Publisher: Nature Publishing Group.

\bibitem[{Trudgill(1986)}]{trudgillDialectsContact1986}
Peter Trudgill. 1986.
\newblock \emph{Dialects in Contact}.
\newblock Number~10 in Language in Society. {B. Blackwell}, {Oxford, UK ; New
  York, NY, USA}.

\bibitem[{Zannettou et~al.(2018)Zannettou, Caulfield, Blackburn, De~Cristofaro,
  Sirivianos, Stringhini, and Suarez-Tangil}]{Zannettou2018}
Savvas Zannettou, Tristan Caulfield, Jeremy Blackburn, Emiliano De~Cristofaro,
  Michael Sirivianos, Gianluca Stringhini, and Guillermo Suarez-Tangil. 2018.
\newblock \href {https://doi.org/10.1145/3278532.3278550} {On the origins of
  memes by means of fringe web communities}.
\newblock In \emph{Proceedings of the Internet Measurement Conference 2018},
  IMC '18, page 188–202, New York, NY, USA. Association for Computing
  Machinery.

\bibitem[{Zhang et~al.(2021)Zhang, Zhang, Zhang, Yang, and
  Lin}]{zhang-etal-2021-multimet}
Dongyu Zhang, Minghao Zhang, Heting Zhang, Liang Yang, and Hongfei Lin. 2021.
\newblock \href {https://doi.org/10.18653/v1/2021.acl-long.249} {{M}ulti{MET}:
  A multimodal dataset for metaphor understanding}.
\newblock In \emph{Proceedings of the 59th Annual Meeting of the Association
  for Computational Linguistics and the 11th International Joint Conference on
  Natural Language Processing (Volume 1: Long Papers)}, pages 3214--3225,
  Online. Association for Computational Linguistics.

\bibitem[{Zhang et~al.(2017)Zhang, Hamilton, Danescu-Niculescu-Mizil, Jurafsky,
  and Leskovec}]{zhangCommunityIdentityUser2017a}
Justine Zhang, William~L. Hamilton, Cristian Danescu-Niculescu-Mizil, Dan
  Jurafsky, and Jure Leskovec. 2017.
\newblock \href {http://arxiv.org/abs/1705.09665} {Community {Identity} and
  {User} {Engagement} in a {Multi}-{Community} {Landscape}}.
\newblock ArXiv:1705.09665 [physics].

\bibitem[{Zhang(2005)}]{zhangChineseYuppieBeijing2005}
Qing Zhang. 2005.
\newblock \href {https://doi.org/10.1017/S0047404505050153} {A {{Chinese}}
  yuppie in {{Beijing}}: {{Phonological}} variation and the construction of a
  new professional identity}.
\newblock \emph{Language in Society}, 34(3):431--466.

\bibitem[{Zhu and Jurgens(2021{\natexlab{a}})}]{zhu2021idiosyncratic}
Jian Zhu and David Jurgens. 2021{\natexlab{a}}.
\newblock Idiosyncratic but not arbitrary: Learning idiolects in online
  registers reveals distinctive yet consistent individual styles.
\newblock In \emph{Proceedings of the 2021 Conference on Empirical Methods in
  Natural Language Processing}, pages 279--297.

\bibitem[{Zhu and Jurgens(2021{\natexlab{b}})}]{zhu2021structure}
Jian Zhu and David Jurgens. 2021{\natexlab{b}}.
\newblock The structure of online social networks modulates the rate of lexical
  change.
\newblock In \emph{Proceedings of the 2021 Conference of the North American
  Chapter of the Association for Computational Linguistics: Human Language
  Technologies}, pages 2201--2218.

\end{thebibliography}
\bibliographystyle{acl_natbib}

\appendix

\section{Details on visually clustering templates}
\label{sec:visual_details}

\subsection{Preprocessing}

\begin{figure*}[t]
    \centering
    \begin{subfigure}{0.23\linewidth}
        \includegraphics[width=\textwidth]{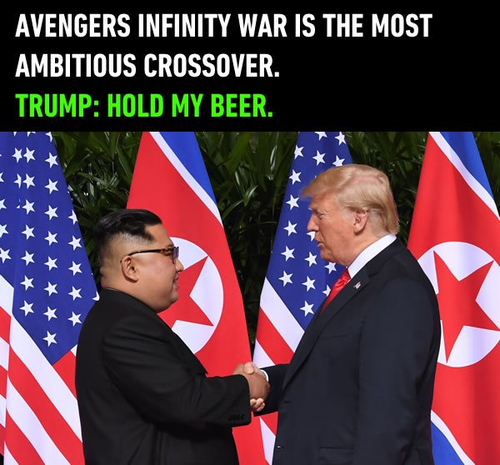}
        \caption{Original image}
        \label{fig:preprocess_examplea}
    \end{subfigure}
    ~
    \begin{subfigure}{0.23\linewidth}
        \centering
        \includegraphics[width=\textwidth]{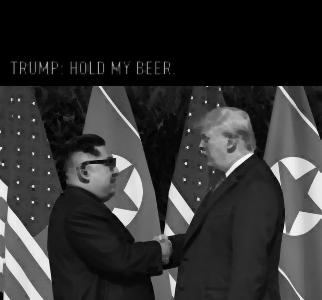}
        \caption{Remove text}
    \end{subfigure}
    ~
    \begin{subfigure}{0.23\linewidth}
        \centering
        \includegraphics[width=\textwidth]{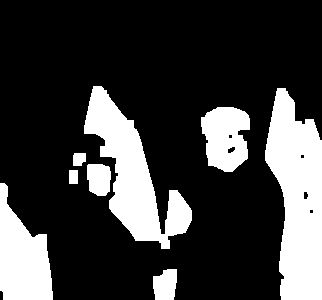}
        \caption{Remove text artifacts}
    \end{subfigure}
    ~
    \begin{subfigure}{0.23\linewidth}
        \centering
        \includegraphics[width=\textwidth]{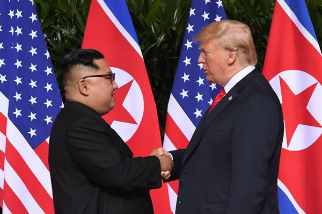}
        \caption{Trim excess borders}
    \end{subfigure}
    
    \caption{Example of the image preprocessing steps, described in Section \ref{templates}}
    \label{fig:preprocess_example}
\end{figure*}

One common meme layout that would caused issues in the template clustering step was a text frame around the source image, where there is a border around a source image, as well as some text above or below (see Figure~\ref{fig:preprocess_examplea} for an example).

For each image, we use a rectangular kernel to detect potential text patches, replace those patches with the background color, and finally identify the bounding box for the remaining source image without any excess borders. Figure~\ref{fig:preprocess_example} walks through the steps visually.

\subsection{Image hashing}

\begin{figure}[h]
    \centering
    \begin{subfigure}{\linewidth}
        \includegraphics[width=1\linewidth]{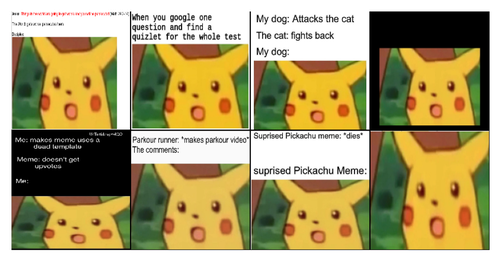}
        \caption{8763e2636178d897}
        \label{fig:phash_a}
    \end{subfigure}

    \begin{subfigure}{\linewidth}
        \includegraphics[width=1\linewidth]{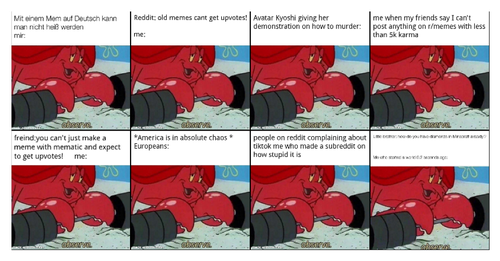}
        \caption{9465a9596e1a66bc}
        \label{fig:phash_b}
    \end{subfigure}
    \caption{Examples of groups of meme instances with the same perceptual hash when processed.}
    \label{fig:phash}
\end{figure}

We create a 64-bit perceptual hash for each preprocessed image in the dataset. The preprocessed images are only used for the hashing step; all other steps use the original image. Figure~\ref{fig:phash} shows examples of images whose preprocessed versions have the same hash. 

\subsection{Hash clustering}

\begin{figure}
    \centering
    \includegraphics[width=1\linewidth]{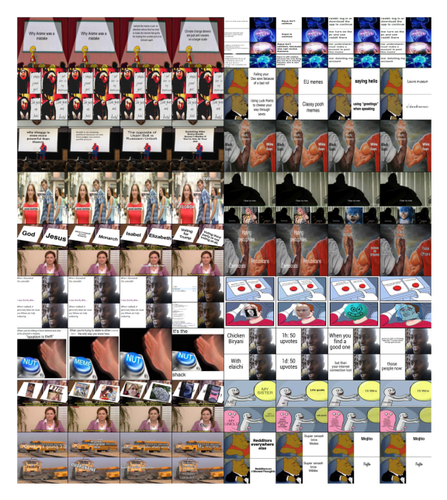}
    \caption{Sample images from the 18 most common perceptual hash clusters; each row contains two clusters with four sampled images.}
    \label{fig:hashclusters}
\end{figure}

We first compute pairwise Hamming distance between all the hashes. Then, we discard any pairs with a Hamming distance greater than 10. Then we construct a network of hashes, where edges of the graph are calculated as $11 - d_{ij}$ for Hamming distance $d_{ij}$ between the $i$th and $j$th hashes before finally using the Leiden algorithm to cluster hashes. Figure~\ref{fig:hashclusters} shows the top 18 most heavily represented hash clusters, with 4 sampled images from each.

We use the Leiden algorithm with the Constant Potts Model (CPM) as the quality function; we use a density of 1.0, but experiments with other density values (0.01, 0.1, 10) yielded qualitatively similar or worse results. The algorithm results in aggressively split clusters, where each cluster is coherent, but there are some memes that share a base template but are split between two clusters (e.g. Winnie the Pooh appears twice in Figure~\ref{fig:hashclusters}, among others).

We find that these duplicate hash clusters are often merged when we generate the semantic clusters.

\section{Details on the semantic clusters}
\label{sec:sem_details}

\subsection{Calculating edge weights.}

To prepare the data for clustering, we constructed a weighted adjacency matrix by keeping only the top 10 nearest neighbors for each template embedding. We calculated the weight as \[
    w_a(b) = \lambda^{r_a(b)}
,\]
where $w_a(b)$ is the weight of the edge between templates $b$ and $a$, $r_a(b)$ is the rank of the cosine similarity between $a$ and $b$, and $\lambda$ is a discount factor (we set this to 0.9).

We chose to weight by a function of ranked similarity instead of cosine similarity directly because we found the cosine similarity was often low even for the embeddings of semantically equivalent templates, resulting poor recall and many small clusters. Using the weighted ranking, we get more templates per cluster without introducing too many false positives.

\subsection{Outputs}

There are 784 semantic clusters generated from the RoBERTa embeddings---table~\ref{tab:cluster_sizes} shows statistics for cluster sizes for all of the models. We include more extensive examples of the clusters in appendix~\ref{sec:sem_cluster_examples}.

\begin{table}
    \centering
    \begin{tabular}{lrr}
        \textbf{Model} & \textbf{\# Clusters} & \textbf{Avg. Size}\\
        \hline
        RoBERTa & 784 & 8.7 \\
        CLIP & 657 & 10.4 \\
        CLIP-diff & 617 & 11.1 \\
        Concat. & 685 & 10.0 \\
    \end{tabular}
    \caption{Count and average sizes of semantic clusters generated from each embedding model.}
    \label{tab:cluster_sizes}
\end{table}

\begin{figure}
    \centering
    \includegraphics[width=1\linewidth]{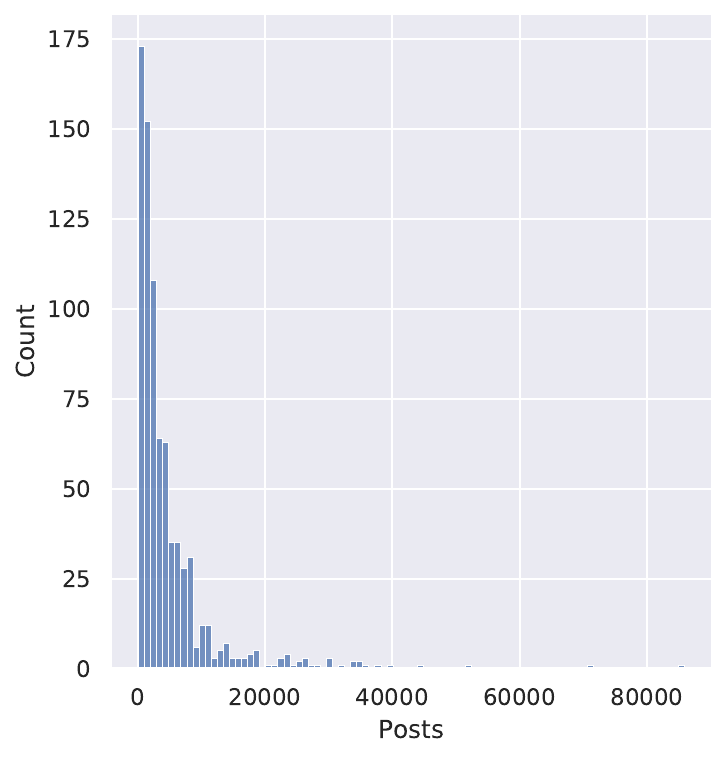}
    \caption{Distribution of semantic cluster coverage}
    \label{fig:sem_cluster_counts}
\end{figure}

The distribution of clusters is highly skewed. Figure~\ref{fig:sem_cluster_counts} shows the distribution of cluster sizes in our Reddit dataset, for the RoBERTa embeddings. The largest 103 clusters account for 50\% of the posts the dataset. The largest 10 account for 12\%.

\section{Model evaluation details}

Human annotators were presented with pairs of images with the following instructions:

\begin{quote}
You will be looking at pairs of memes; for each pair, you will be answering two yes/no questions.
\begin{enumerate}
    \item is this pair semantically similar (can you conceivably copy / paste the text of one into the other with minor changes and have it still make sense)
    \item is this pair visually similar (do they have the same characters, art style, etc? e.g. two harry potter memes. If the layout is the same but the characters are different, you should mark it as not visually similar)
\end{enumerate}
\end{quote}

\begin{figure}
    \centering
    \begin{subfigure}{\linewidth}
        \includegraphics[width=\textwidth]{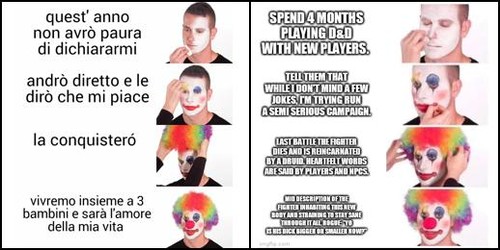}
        \caption{Semantically and visually similar}
    \end{subfigure}
    \begin{subfigure}{\linewidth}
        \includegraphics[width=\textwidth]{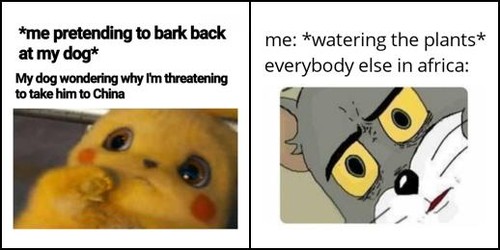}
        \caption{Semantically similar, visually different}
    \end{subfigure}
    \begin{subfigure}{\linewidth}
        \includegraphics[width=\textwidth]{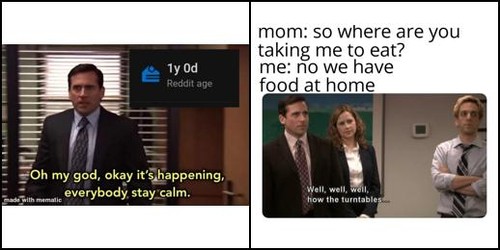}
        \caption{Semantically different, visually similar}
    \end{subfigure}
    \begin{subfigure}{\linewidth}
        \includegraphics[width=\textwidth]{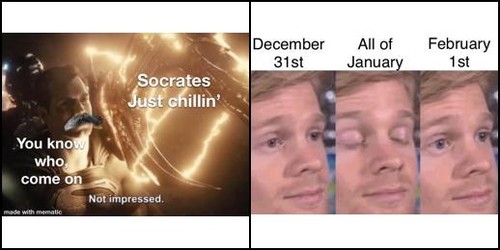}
        \caption{Semantically and visually different}
    \end{subfigure}
    \caption{Example meme pairs for annotation}
    \label{fig:example_pairs}
\end{figure}

Figure~\ref{fig:example_pairs} includes examples of some of the image pairs presented to annotators, as well as the expected judgments for semantic and visual similarity.

We calculate metrics on the model judgments over the set of all annotated pairs (across models). This not only allows us to evaluate on a larger set of annotations, but also helps highlight differences between models.

\begin{figure}
    \centering
    \includegraphics[width=\linewidth]{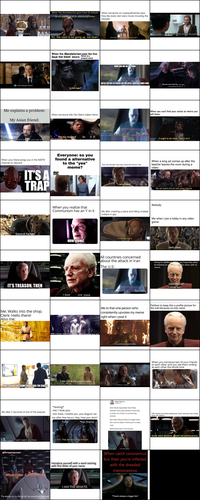}
    \caption{This Concat cluster has a low visual-adjusted precision. }
    \label{fig:visual_bias}
\end{figure}

We found that CLIP-based methods would include clusters that were visually similar but semantically different. Figure~\ref{fig:visual_bias} shows an example of a cluster generated with CLIP embeddings that contains meme templates that have different semantic functions, but almost all contain characters from the \textit{Star Wars} franchise. This cluster had an visually adjusted precision score of -2.08.

\begin{figure*}
    \centering
    \begin{subfigure}{0.4\textwidth}
        \centering
        \includegraphics[width=\linewidth]{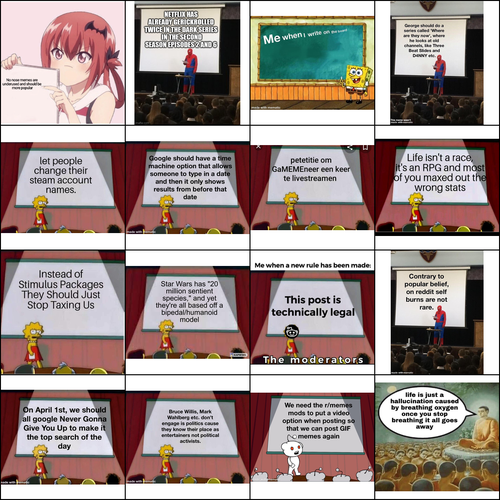}
        \caption{RoBERTa Cluster 0}
    \end{subfigure}
    \begin{subfigure}{0.4\textwidth}
        \centering
        \includegraphics[width=\linewidth]{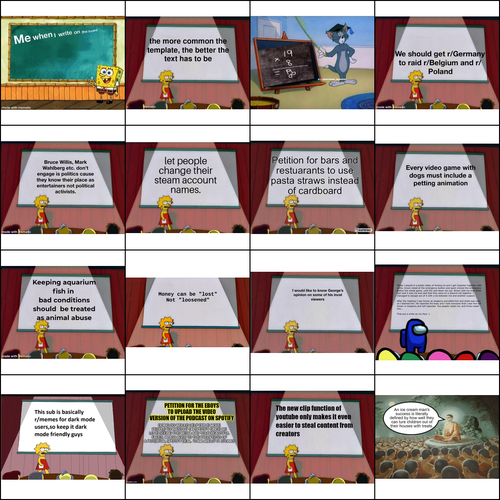}
        \caption{Concat Cluster 2}
    \end{subfigure}
    \begin{subfigure}{0.4\textwidth}
        \centering
        \includegraphics[width=\linewidth]{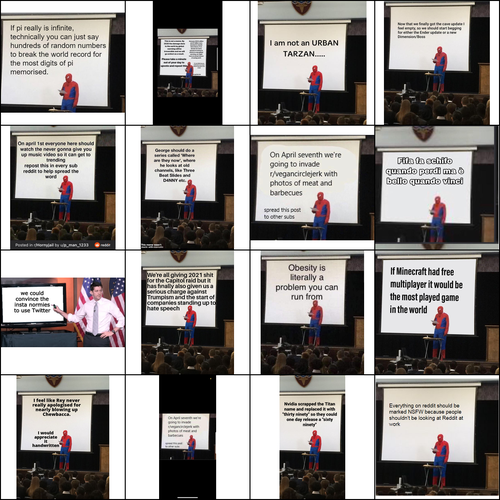}
        \caption{Concat Cluster 31}
    \end{subfigure}
    \begin{subfigure}{0.4\textwidth}
        \centering
        \includegraphics[width=\linewidth]{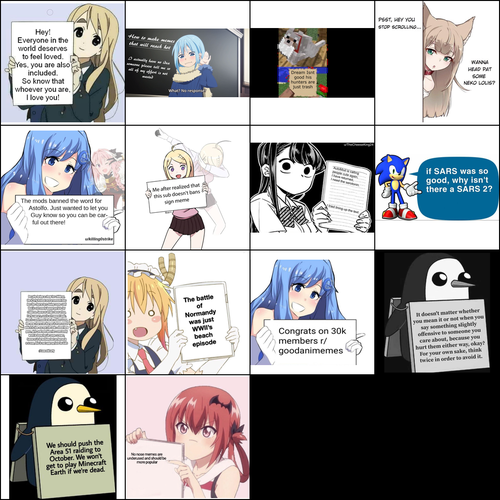}
        \caption{Concat Cluster 134}
    \end{subfigure}
    \caption{Templates that are clustered together by RoBERTa appear in stylistically delineated semantic clusters in the Concat clusters. Displayed are a sample of up to 16 templates from each cluster.}
    \label{fig:stylesplit}
\end{figure*}

The failure case of over-indexing on visual similarity is not restricted to creating semantically incoherent clusters with visual similarity. The CLIP-diff and, to a greater extent, Concat embeddings were good at surfacing less-used variants of templates. However, they do so at the expense of splitting into several stylistically delineated semantic clusters, which is detrimental to our desired analysis on variation. Figure~\ref{fig:stylesplit} shows how templates from the large RoBERTa cluster for declarative templates are divided into several Concat clusters.

\section{Further semantic cluster examples}
\label{sec:sem_cluster_examples}

\begin{figure*}
    \begin{subfigure}{0.5\linewidth}
        \centering
        \includegraphics[width=\textwidth]{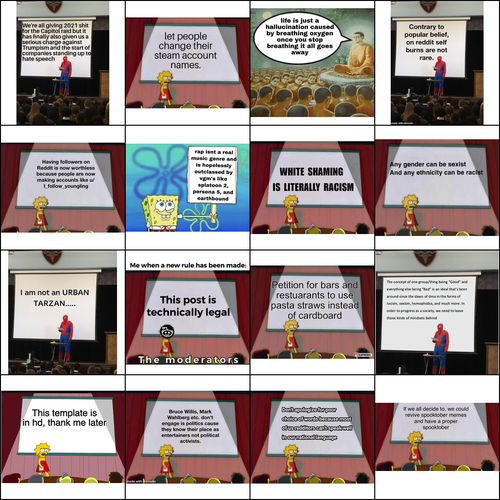}
        \caption{Cluster 0}
    \end{subfigure}~
    \begin{subfigure}{0.5\linewidth}
        \centering
        \includegraphics[width=\textwidth]{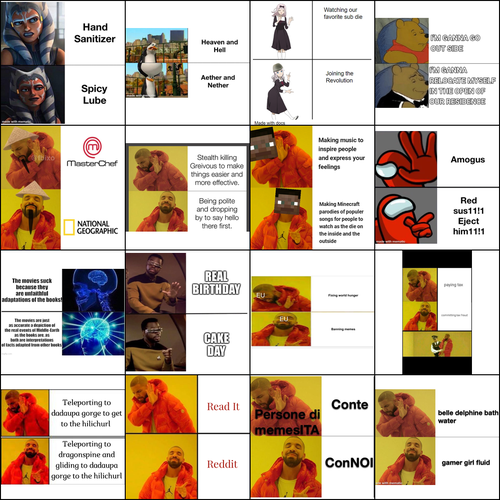}
        \caption{Cluster 1}
    \end{subfigure}
    \begin{subfigure}{0.5\linewidth}
        \centering
        \includegraphics[width=\textwidth]{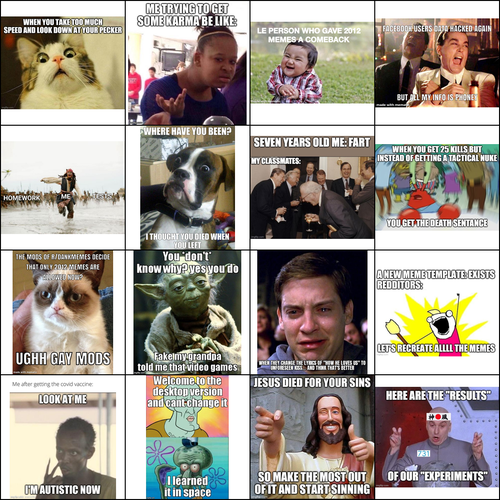}
        \caption{Cluster 10}
    \end{subfigure}~
    \begin{subfigure}{0.5\linewidth}
        \centering
        \includegraphics[width=\textwidth]{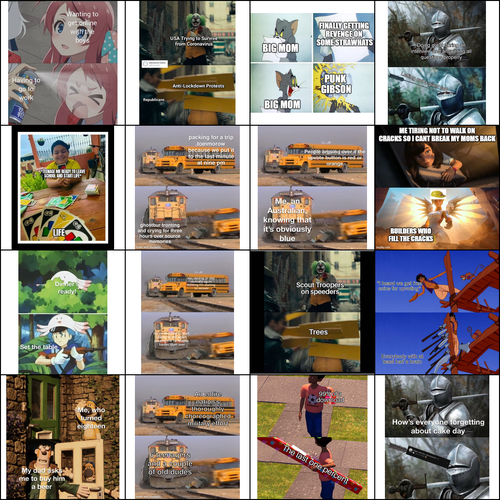}
        \caption{Cluster 30}
    \end{subfigure}
    \caption{Samples from RoBERTa clusters}
    \label{fig:roberta_sem}
\end{figure*}

\begin{figure*}
    \begin{subfigure}{0.5\linewidth}
        \centering
        \includegraphics[width=\textwidth]{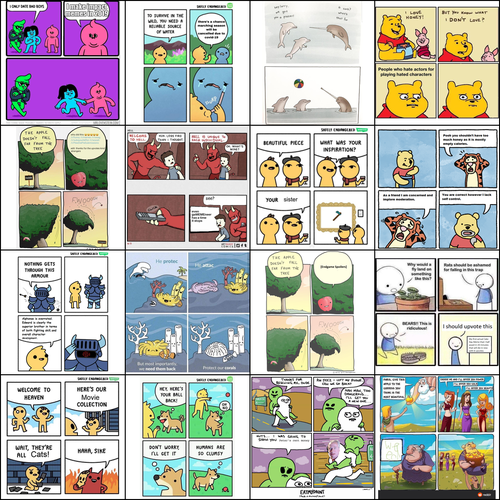}
        \caption{Cluster 0}
    \end{subfigure}~
    \begin{subfigure}{0.5\linewidth}
        \centering
        \includegraphics[width=\textwidth]{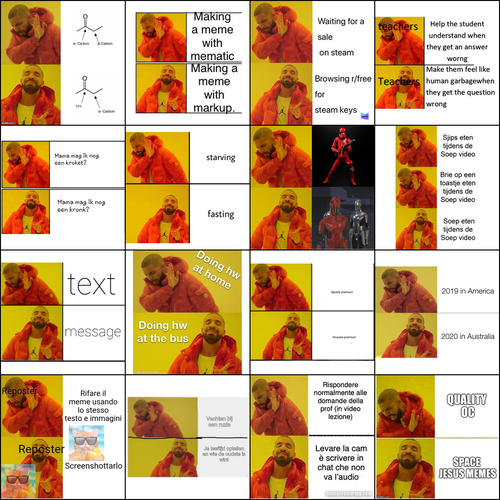}
        \caption{Cluster 1}
    \end{subfigure}
    \begin{subfigure}{0.5\linewidth}
        \centering
        \includegraphics[width=\textwidth]{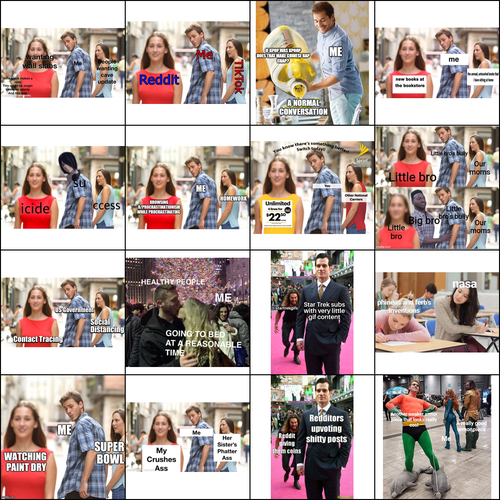}
        \caption{Cluster 36}
    \end{subfigure}~
    \begin{subfigure}{0.5\linewidth}
        \centering
        \includegraphics[width=\textwidth]{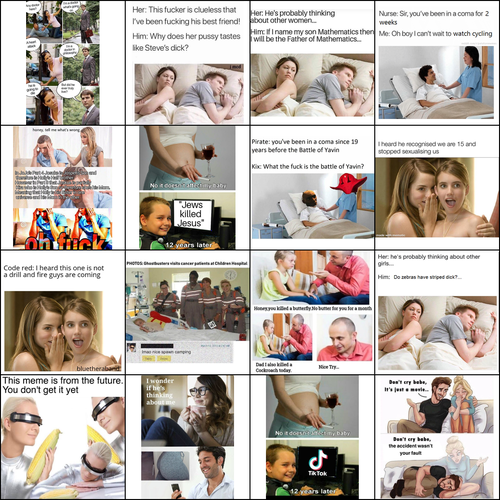}
        \caption{Cluster 23}
    \end{subfigure}
    \caption{Samples from CLIP clusters}
    \label{fig:clip_sem}
\end{figure*}

\begin{figure*}
    \begin{subfigure}{0.5\linewidth}
        \centering
        \includegraphics[width=\textwidth]{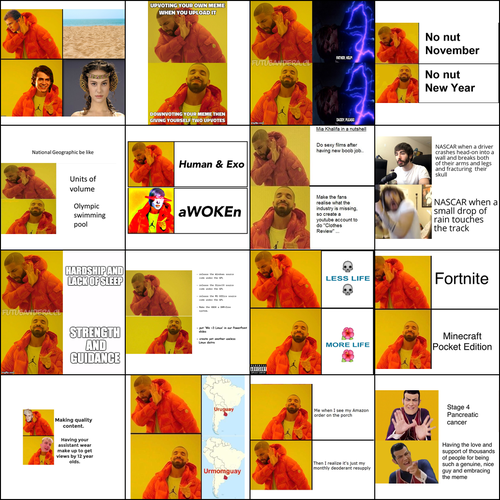}
        \caption{Cluster 0}
    \end{subfigure}~
    \begin{subfigure}{0.5\linewidth}
        \centering
        \includegraphics[width=\textwidth]{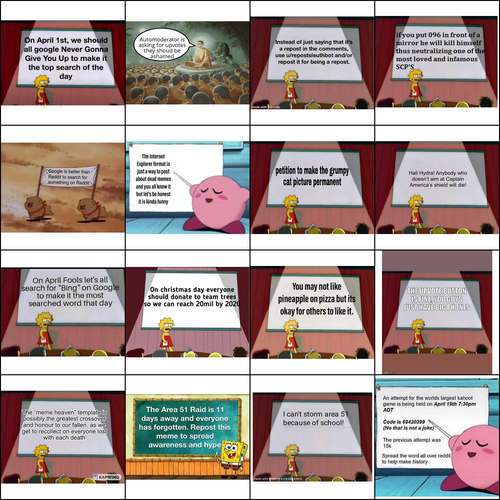}
        \caption{Cluster 1}
    \end{subfigure}
    \begin{subfigure}{0.5\linewidth}
        \centering
        \includegraphics[width=\textwidth]{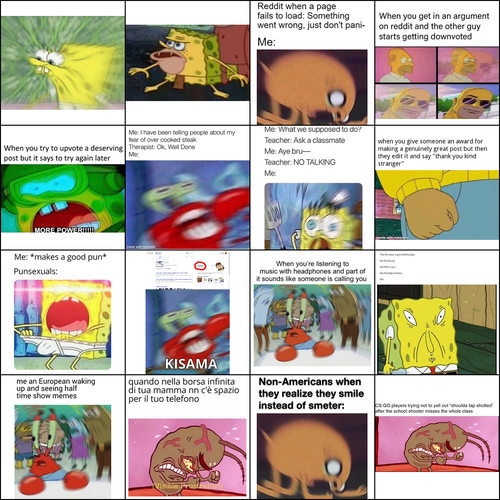}
        \caption{32}
    \end{subfigure}~
    \begin{subfigure}{0.5\linewidth}
        \centering
        \includegraphics[width=\textwidth]{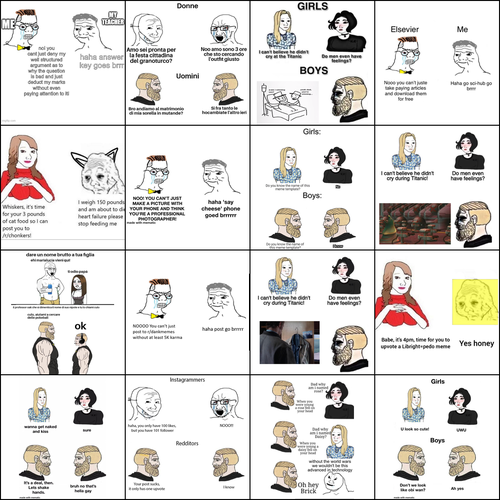}
        \caption{Cluster 22}
    \end{subfigure}
    \caption{Samples from CLIP-diff clusters}
    \label{fig:clipdiff_sem}
\end{figure*}

\begin{figure*}
    \begin{subfigure}{0.5\linewidth}
        \centering
        \includegraphics[width=\textwidth]{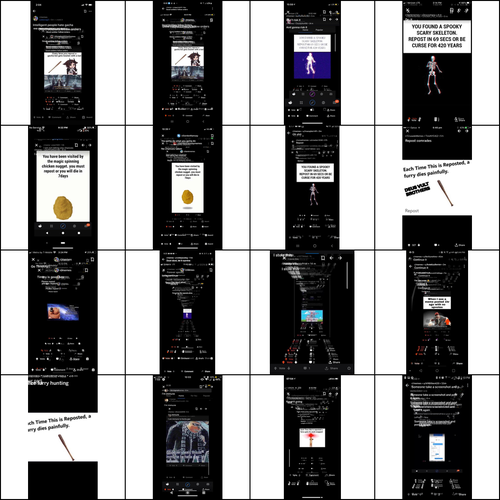}
        \caption{Cluster 0}
    \end{subfigure}~
    \begin{subfigure}{0.5\linewidth}
        \centering
        \includegraphics[width=\textwidth]{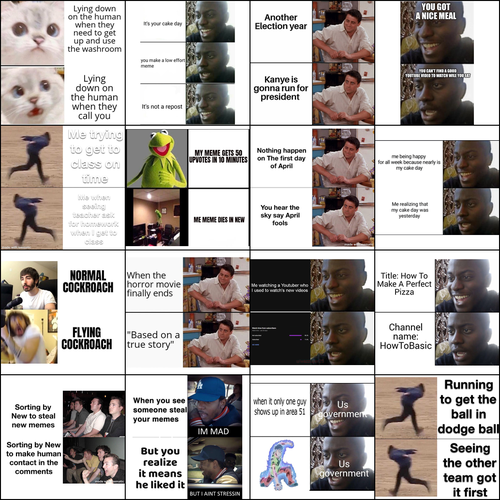}
        \caption{Cluster 1}
    \end{subfigure}
    \begin{subfigure}{0.5\linewidth}
        \centering
        \includegraphics[width=\textwidth]{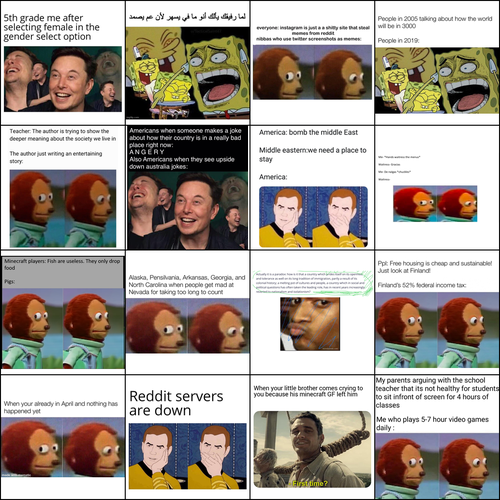}
        \caption{Cluster 15}
    \end{subfigure}~
    \begin{subfigure}{0.5\linewidth}
        \centering
        \includegraphics[width=\textwidth]{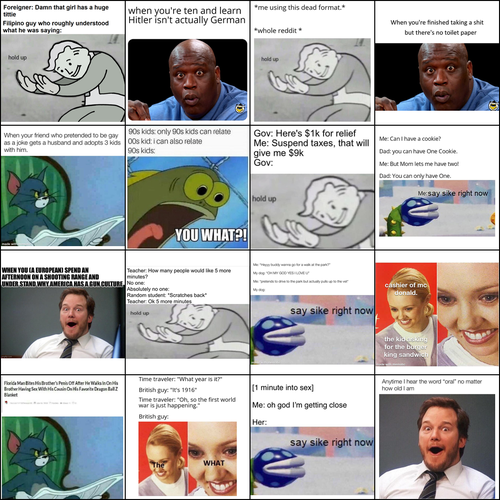}
        \caption{Cluster 29}
    \end{subfigure}
    \caption{Samples from CLIP-diff + RoBERTa clusters}
    \label{fig:concat_sem}
\end{figure*}

Figures~\ref{fig:roberta_sem},\ref{fig:clip_sem},\ref{fig:clipdiff_sem},\ref{fig:concat_sem} contain more examples of semantic clusters generated using the different embedding models.

\end{document}